\documentclass[11pt]{article}

\usepackage[preprint]{acl}

\usepackage{times}
\usepackage{latexsym}

\usepackage[T1]{fontenc}

\usepackage[utf8]{inputenc}

\usepackage{microtype}

\usepackage{inconsolata}
\usepackage{makecell} 

\usepackage{graphicx}
\usepackage{amsmath}
\usepackage{amsfonts}
\usepackage{booktabs} 
\usepackage{multirow}
\newtheorem{theorem}{Theorem}
\newtheorem{lemma}{Lemma}
\newtheorem{definition}{Definition}
\usepackage{algorithm}
\usepackage{algpseudocode}
\usepackage{amssymb}
\usepackage{bm}

\title{Toward Consistent World Models with Multi-Token Prediction and Latent Semantic Enhancement}

\author{
  Qimin Zhong$^{1}$, Hao Liao$^{1}$, Haiming Qin$^{1}$, Mingyang Zhou$^{1}$, \\
  \textbf{Rui Mao$^{1}$, Wei Chen$^{2}$, Naipeng Chao$^{1}$} \\
  $^{1}$Shenzhen University, $^{2}$Microsoft Research Asia \\
    \texttt{\{2023044007@email, haoliao, 2453103002@mails, zmy, mao, npchao\}@szu.edu.cn} \\
  \texttt{weic@microsoft.com}
}

\begin{document}
\maketitle
\begin{abstract}
Whether Large Language Models (LLMs) develop coherent internal world models remains a core debate. While conventional Next-Token Prediction (NTP) focuses on one-step-ahead supervision,
Multi-Token Prediction (MTP) has shown promise in learning more structured representations. In this work, we provide a theoretical perspective analyzing the gradient inductive bias of MTP, supported by empirical evidence, showing that MTP promotes the convergence toward internal belief states by inducing representational contractivity via gradient coupling. However, we reveal that standard MTP often suffers from structural hallucinations, where discrete token supervision encourages illegal shortcuts in latent space that violate environmental constraints. To address this, we propose a novel method \textbf{Latent Semantic Enhancement MTP (LSE-MTP)}, which anchors predictions to ground-truth hidden state trajectories. Experiments on synthetic graphs and real-world Manhattan Taxi Ride show that LSE-MTP effectively bridges the gap between discrete tokens and continuous state representations, enhancing representation alignment, reducing structural hallucinations, and improving robustness to perturbations.
\end{abstract}

\section{Introduction}
\label{sec:introduction}

Internalizing the dynamics of an environment is a hallmark of intelligent behavior. This capability, often formalized as a world model~\citep{ha2018world, schmidhuber1990making}, allows an agent to reason beyond immediate observations by simulating how states evolve over time~\citep{silver2017predictron, schrittwieser2020mastering}. Rather than reacting myopically to inputs, systems equipped with world models can anticipate future outcomes, evaluate alternative trajectories, and plan accordingly. The success of DreamerV3~\citep{hafner2025mastering} vividly illustrates how learning internal dynamics can yield strong generalization across diverse tasks, even under limited supervision.
Recent evidence suggests that the fidelity of internal world models is a key driver of post‑training potential and correlates with improved reasoning and downstream performance~\citep{zhai2024better}.

In the context of Natural Language Processing, this perspective raises a fundamental and intriguing question: do Large Language Models (LLMs) trained purely through Next-Token Prediction (NTP) develop meaningful internal world models~\citep{brown2020language, rae2021scaling}? While NTP has proven remarkably effective at scaling language understanding and generation, its optimization objective is inherently local, as it primarily focuses on predicting the likelihood of the next symbol given a context. As a result, such models often excel at capturing surface-level regularities but struggle to consistently internalize deeper global structure or long-range dynamics, especially when complex reasoning requires maintaining coherent latent states over extended horizons~\citep{bachmann2024the, wyatt2025alternatives}.

This concern has been substantiated by recent real-world evaluations.
\citet{vafa2024evaluating} introduce a world-model benchmark based on Manhattan taxi trajectories,
where city streets are abstracted as a graph with explicit topological constraints.
Despite achieving near-perfect next-step prediction accuracy,
NTP-trained models frequently fail to encode the global structure of the street network in their latent states,
leading to invalid routes and severe fragility under minor perturbations.
These findings demonstrate that strong token-level performance alone does not guarantee a coherent internal world model.

Multi-Token Prediction (MTP) has recently emerged as a promising alternative~\citep{gloeckle2024better}. By supervising multiple future tokens simultaneously, MTP encourages models to look beyond immediate continuations and consider longer-term evolution. This shift in supervision fundamentally alters the training signal: instead of fitting isolated conditional distributions, the model is pressured to represent how sequences unfold over time. From a representation-learning standpoint, such foresight can induce representational contractivity, encouraging diverse historical contexts to converge toward shared internal belief states that summarize the underlying environment. This phenomenon suggests a potential pathway for LLMs to move from shallow sequence modeling toward more structured internal representations resembling world models.

Yet, the presence of foresight alone does not guarantee coherent internal reasoning. In practice, we observe that MTP-trained models can develop a subtle but systematic failure mode, which we refer to as structural hallucination. Even when long-term predictions are accurate at the token level, the latent evolution that supports them may violate essential constraints of the environment. Intermediate steps can be implicitly skipped, transitions may become implausible, and internal trajectories can exploit shortcuts that would be invalid under the true dynamics. This reveals a key tension: optimizing distant predictions without explicit trajectory-level grounding can incentivize models to prioritize outcomes over the integrity of the underlying process.

These observations point to a broader gap between discrete supervision and continuous internal dynamics. Token-level objectives, even when extended to multiple future steps, offer limited control over how representations evolve over time. In the absence of mechanisms that explicitly align latent transitions with valid state progressions, models may develop internally inconsistent simulations that appear accurate only at their final predictions. Bridging this gap is crucial for elevating multi-token prediction from a stronger forecasting objective to a dependable foundation for world modeling and long-horizon planning.

This work relates to several active research threads, including world models in language modeling, multi-token prediction, latent state consistency, and graph-based planning. Detailed discussion is deferred to Appendix~\ref{app:related-works}.

To summarize, our main contributions are highlighted by the following three perspectives:

\begin{itemize}
    \item We provide a theoretical analysis of the gradient coupling mechanism in Multi-Token Prediction (MTP), showing how it induces contractivity that facilitates the emergence of belief states, while exposing a structural hallucination risk arising from overemphasis on distant targets over local connectivity.
    
    \item We propose LSE-MTP, a framework that enforces latent consistency by aligning multi-token predictions with ground-truth hidden state trajectories and semantic anchors, thereby enforcing valid stepwise transitions and discouraging illegal shortcuts.
    
    \item Through extensive experiments on synthetic graphs and real-world Manhattan taxi navigation, we show that LSE-MTP improves path legality, belief compression, and robustness to perturbations in multi-step planning.
\end{itemize}

\begin{figure*}[t]
    \centering
    \includegraphics[width=0.95\textwidth]{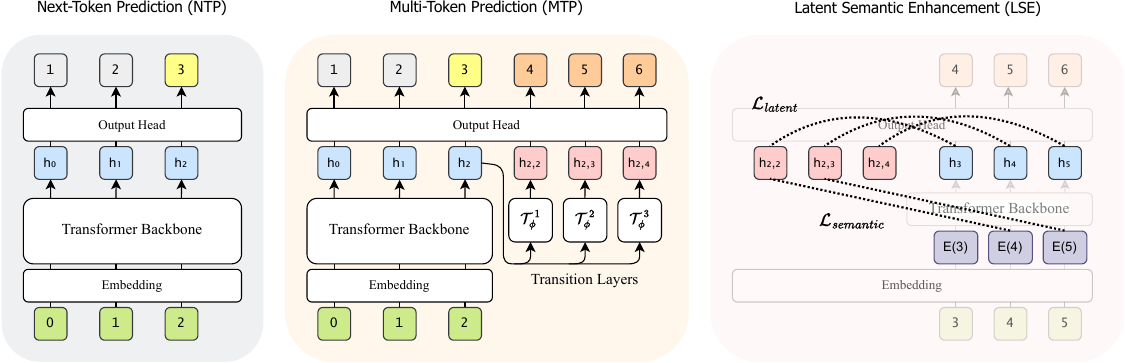}
    \caption{
    \textbf{Overview of LSE-MTP.}
    Given a backbone hidden state $\mathbf{h}_n$, horizon-specific transition layers
    produce multi-step predictive representations.
    Training combines multi-step token prediction with latent consistency
    and semantic anchoring losses.
    All transition layers are discarded at inference time.
    }
    \label{fig:lse_architecture}
\end{figure*}

\section{Preliminaries}
\label{sec:preliminaries}

\subsection{Next-Token Prediction}

The standard paradigm for autoregressive sequence modeling is Next-Token Prediction (NTP).
Given a history $H_n = (u_1, \dots, u_n)$, the objective minimizes the negative log-likelihood of the next token:
\begin{equation}
    \mathcal{L}_{\text{NTP}}(\theta) = 
    \mathbb{E}_{S \sim \mathcal{D}, n} 
    \big[ -\log P_\theta(u_{n+1} \mid H_n) \big].
\end{equation}

Despite its empirical success, NTP exhibits limitations in structured reasoning tasks:
(i) it primarily fits local co-occurrence statistics rather than invariant transition rules~\citep{wu2024can},
and (ii) under teacher forcing, models can exploit local token correlations to bypass global reasoning, leading to the acquisition of shortcuts during training that fail to generalize to the underlying task logic~\citep{bachmann2024the, arvid2025language}.

\subsection{Multi-Token Prediction}

Multi-Token Prediction (MTP) extends NTP by jointly predicting the next $K$ future tokens during training,
while retaining standard autoregressive decoding at inference~\citep{gloeckle2024better}.

We consider an MTP architecture with a shared output head and horizon-specific transition layers.
Given the backbone hidden state $\mathbf{h}_n = f_\theta(H_n)$,
the next token $u_{n+1}$ is predicted directly, while the $k$-step future token ($k \ge 2$)
is predicted from a transformed representation $\mathcal{T}_\phi^{(k-1)}(\mathbf{h}_n)$.
All predictions for different horizons are decoded by the same shared output head.
The training objective is:
\begin{equation}
\mathcal{L}_{\text{MTP}} =
\mathbb{E}_{S,n}
\Big[
\mathcal{L}^{(1)}(\mathbf{h}_n, u_{n+1})
+
\sum_{k=2}^{K}
\mathcal{L}^{(k)}(\mathbf{h}_n, u_{n+k})
\Big].
\end{equation}

\subsection{Representation Space and Belief States}

The hidden state $\mathbf{h}_n$ serves as a compact summary of the history $H_n$
and implicitly encodes information about future trajectories.

\begin{definition}
The set of hidden states $\mathcal{H} = \{\mathbf{h}_n\}$ forms a \textbf{representation space},
where histories with similar future continuations are embedded nearby~\citep{littman2001predictive}.
\end{definition}

\begin{definition}
The idealized representation associated with $\mathbf{h}_n$ is a \textbf{belief state} $\mathbf{b}_n$~\citep{kaelbling1998planning},
satisfying
\begin{equation}
P(u_{n+1:\infty} \mid H_n) \approx P(u_{n+1:\infty} \mid \mathbf{b}_n).
\end{equation}
Belief states provide a compact internal model of future dynamics.
\end{definition}

\section{A Theoretical Perspective on Multi-Token Prediction}
\label{sec:theory}

Before diving into mathematical analysis, we provide the intuition behind MTP's impact. By predicting multiple tokens simultaneously, MTP encourages histories leading to the same future to "merge" within the representation space. This merging is inherently blind: it constrains only future outcomes while ignoring intermediate states, which can produce illegal shortcuts in latent space. In this section, we formally characterize this behavior using gradient flow dynamics.

To obtain a tractable analytic framework, we focus on the \textbf{linearized regime} (lazy training), approximating the optimization trajectory via the local Neural Tangent Kernel (NTK)~\citep{chizat2019lazy}. This local linearization captures the instantaneous directional pressure exerted by the loss on the representation space.

Let $\mathbf{h} = f_\theta(H)$ denote the hidden state of a backbone parameterized by $\theta$, evolving under gradient flow $\dot{\theta} = -\eta \nabla_\theta \mathcal{L}$. We define the representation-level NTK as:
\[
\mathbf{K}(\mathbf{h}_i, \mathbf{h}_j) = \nabla_\theta f_\theta(H_i) \nabla_\theta f_\theta(H_j)^\top \in \mathbb{R}^{d \times d}.
\]

\begin{definition}
\label{def:equivalence}
Two hidden states $\mathbf{h}_1$ and $\mathbf{h}_2$ are \textbf{$\textbf{k}$-step future equivalent} ($\mathbf{h}_1 \sim_k \mathbf{h}_2$) if they are supervised by the same $k$-step-ahead target token $y^* = u_{n+k} = u_{m+k}$ under the $(k\!-\!1)$-th transition layer $\mathcal{T}^{(k-1)}$ and the shared prediction head.
\end{definition}

\begin{definition}
\label{def:contractivity}
The representation space exhibits \textbf{contractivity} for a pair of histories if the time derivative of the squared distance $\mathcal{D}(\mathbf{h}_1, \mathbf{h}_2) = \| \mathbf{h}_1 - \mathbf{h}_2 \|^2$ satisfies $\dot{\mathcal{D}} \le 0$ under gradient flow, indicating convergence toward a unified belief state.
\end{definition}

Based on these definitions, we compare the geometric effects of NTP and MTP. Formal derivations are deferred to Appendix~\ref{app:proofs}.

\begin{theorem}
\label{thm:ntp}
Under the NTP loss $\mathcal{L}_{\text{NTP}}$, the contractive condition $\dot{\mathcal{D}} \le 0$ holds primarily for $1$-step equivalent states ($\mathbf{h}_1 \sim_1 \mathbf{h}_2$). For states with different next-step targets, the gradients $\nabla_{\mathbf{h}} \mathcal{L}$ tend to point in opposite directions, preserving representational separation.
\end{theorem}

\begin{theorem}
\label{thm:mtp}
Under the MTP loss $\mathcal{L}_{\text{MTP}}$, consider $k$-step future-equivalent states $\mathbf{h}_1 \sim_k \mathbf{h}_2$ with different immediate targets $u_{n+1} \neq u_{m+1}$. A $k$-step update on $\mathbf{h}_1$ induces a positive cross-update on the corresponding logit of $\mathbf{h}_2$, $\dot{z}_{y_1}(\mathbf{h}_2) > 0$, where the gradients $\nabla_{\mathbf{h}_1} \mathcal{L}^{(k)}_1$ and $\nabla_{\mathbf{h}_2} \mathcal{L}^{(k)}_1$ align through the cross-history NTK $\mathbf{K}(\mathbf{h}_1, \mathbf{h}_2)$, facilitating a predictive coupling that can partially blur the representational separation between distinct trajectories.
\end{theorem}

\textbf{Intuition:} If two histories share an identical future, training on one trajectory inadvertently increases the prediction confidence of the other's next token, even if their immediate targets differ.

\begin{lemma}
\label{lem:mtp}
For a pair of $k$-step future-equivalent states ($\mathbf{h}_1 \sim_k \mathbf{h}_2$), 
a full-rank transition Jacobian ensures that MTP induces a stable contractive force 
with $\dot{\mathcal{D}} \le 0$. 
The resulting geometric flow is governed by $\mathbf{K}\mathbf{S}$, 
where $\mathbf{K}$ is the NTK and $\mathbf{S}$ the pull-back Hessian. 
Although $\mathbf{K}\mathbf{S}$ is generally non-symmetric, it is similar to a symmetric PSD matrix, 
implying real, non-negative eigenvalues and thus local convergence to a unified belief state.
\end{lemma}

\textbf{Intuition:} This predictive coupling manifests as a geometric force that pulls together the representations of different histories whenever they lead to a common future.

These results demonstrate that MTP induces geometric contraction among representations sharing future dynamics. This effect facilitates the alignment of future-equivalent states (Section~\ref{sec:5.2.1}) and the compression of diverse histories into unified belief representations (Section~\ref{sec:5.2.2}). However, the contraction is inherently \textit{outcome-driven}, ignoring the physical validity of intermediate transitions. As shown in our linear model (Section~\ref{sec:5.1}), MTP can induce transition weights toward unobserved states that happen to lead to the same target. 

This phenomenon leads to \textit{structural hallucinations}, where probability mass is incorrectly assigned to illegal shortcuts in latent space, causing the model to deviate from the true trajectory (Section~\ref{sec:5.2.3}). This theoretical gap motivates the development of our \textbf{LSE-MTP} framework (Section~\ref{sec:method}), which effectively anchors MTP-induced contraction to ground-truth latent trajectories.

\section{What is LSE-MTP}
\label{sec:method}

We introduce \textbf{Latent Semantic Enhancement (LSE)}, a training framework built on
Multi-Token Prediction (MTP) with a prediction horizon of $K$.
Given backbone hidden states $\mathbf{h}_n \in \mathbb{R}^d$,
we employ horizon-specific transition layers
$\{\mathcal{T}_\phi^{(k-1)}\}_{k=2}^K$
to produce $k$-step predictive representations
$\hat{\mathbf{h}}_{n,k} = \mathcal{T}_\phi^{(k-1)}(\mathbf{h}_n)$,
with $\hat{\mathbf{h}}_{n,1} = \mathbf{h}_n$.

The training objective consists of three components.
First, we apply a multi-step cross-entropy loss:
\begin{equation}
\mathcal{L}_{ce}
=
\sum_{k=1}^K
\mathbb{E}_n
\big[
-\log P(u_{n+k} \mid \hat{\mathbf{h}}_{n,k})
\big].
\end{equation}

Second, a \textbf{latent consistency loss} aligns predictive representations
with future backbone states:
\begin{equation}
\mathcal{L}_{latent}
=
\sum_{k=2}^K
\mathbb{E}_n
\big\|
\hat{\mathbf{h}}_{n,k}
-
\mathbf{h}_{n+k-1}
\big\|_2^2 .
\end{equation}

Third, a \textbf{semantic anchoring loss} aligns predictive representations with the target token embeddings $\mathbf{E}(\cdot)$:
\begin{equation}
\mathcal{L}_{semantic} = \sum_{k=2}^K \mathbb{E}_n \big\| \hat{\mathbf{h}}_{n,k} - \mathrm{sg}\big(\mathbf{E}(u_{n+k})\big) \big\|_2^2 ,
\end{equation}
where $\mathrm{sg}(\cdot)$ denotes the stop-gradient operator, 
and $\mathbf{E}(\cdot)$ denotes the model's embedding layer.

The full training objective is:
\begin{equation}
\mathcal{L}_{total}
=
\mathcal{L}_{ce}
+
\lambda_l \mathcal{L}_{latent}
+
\lambda_s \mathcal{L}_{semantic}.
\end{equation}
Unless otherwise specified, we set $\lambda_l = \lambda_s = 0.1$.

At inference time, all transition layers and auxiliary losses are discarded,
and decoding follows standard autoregressive NTP.
The complete architecture of the model is illustrated in Figure~\ref{fig:lse_architecture}.

\begin{table*}[htbp]
\centering
\caption{\textbf{Representation alignment on ER and USG graphs.} Sim(F) and Gain denote cosine similarity and structure gain (Sim(F) - random baseline) for $k$-step future equivalent states.}
\label{tab:exp1}
\small
\setlength{\tabcolsep}{5pt}
\begin{tabular}{l | cc | cc | cc || cc | cc | cc}
\toprule
\multirow{2}{*}{\textbf{Model}} & \multicolumn{6}{c||}{\textbf{ER (Erdős--Rényi Graph)}} & \multicolumn{6}{c}{\textbf{USG (Urban Street Graph)}} \\
\cmidrule{2-13}
& \multicolumn{2}{c|}{$k=2$} & \multicolumn{2}{c|}{$k=3$} & \multicolumn{2}{c||}{$k=4$} & \multicolumn{2}{c|}{$k=2$} & \multicolumn{2}{c|}{$k=3$} & \multicolumn{2}{c}{$k=4$} \\
\cmidrule{2-13}
& Sim(F) & Gain & Sim(F) & Gain & Sim(F) & Gain & Sim(F) & Gain & Sim(F) & Gain & Sim(F) & Gain \\
\midrule
1TP & 0.051 & 0.027 & 0.054 & 0.022 & 0.078 & 0.036 & 0.055 & -0.005 & 0.082 & 0.018 & 0.072 & 0.005 \\
2TP & 0.232 & \textbf{0.210} & 0.102 & 0.074 & 0.094 & 0.062 & 0.264 & \textbf{0.214} & 0.126 & 0.066 & 0.112 & 0.048 \\
3TP & 0.229 & 0.195 & 0.194 & \textbf{0.167} & 0.136 & 0.107 & 0.249 & 0.197 & 0.244 & \textbf{0.186} & 0.148 & 0.083 \\
4TP & 0.223 & 0.176 & 0.201 & 0.162 & 0.204 & \textbf{0.171} & 0.230 & 0.178 & 0.235 & 0.180 & 0.222 & \textbf{0.163} \\
\bottomrule
\end{tabular}
\end{table*}

\section{Understanding Multi-Token Prediction in Modeling}
\label{sec:experiments}

In this section, we present two progressive experiments
to empirically examine the theoretical analysis of
Multi-Token Prediction (MTP) developed in Section~\ref{sec:theory}.

\subsection{How Multi-Token Prediction Induces Gradient Coupling}
\label{sec:5.1}

\begin{figure}[b]
    \centering
    \includegraphics[width=0.4\columnwidth]{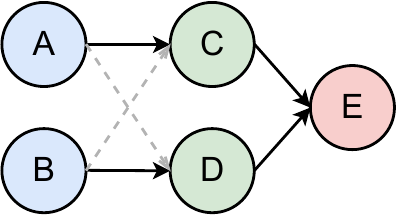}
    \caption{Two independent paths ($A \to C \to E$ and $B \to D \to E$) converging at a shared future $E$.}
    \label{fig:linear_setup}
\end{figure}

To isolate the gradient coupling mechanism of MTP from nonlinear confounders, we construct a minimal linear model. The states $\{A,B,C,D,E\}$ are represented as orthogonal basis vectors in $\mathbb{R}^5$, enabling a transparent analysis of how multi-step supervision reshapes local transition structure. The model has two learnable parameters: a backbone matrix $\bm{W}^B$ for one-step prediction ($\mathbf{h}_{t+1}=\bm{W}^B\mathbf{h}_t$) and, in the 2TP setting, an additional transition matrix $\bm{W}^T$ for predicting the state two steps ahead ($\mathbf{h}_{t+2}=\bm{W}^T\mathbf{h}_{t+1}$).

The task contains two trajectories, $A \to C \to E$ and $B \to D \to E$ (Figure~\ref{fig:linear_setup}). We compare one-token prediction (1TP), optimizing only $\bm{W}^B$, with two-token prediction (2TP), jointly optimizing $\bm{W}^B$ and $\bm{W}^T$, with uniform initialization.

\begin{figure}[t]
    \centering
    \includegraphics[width=0.85\columnwidth]{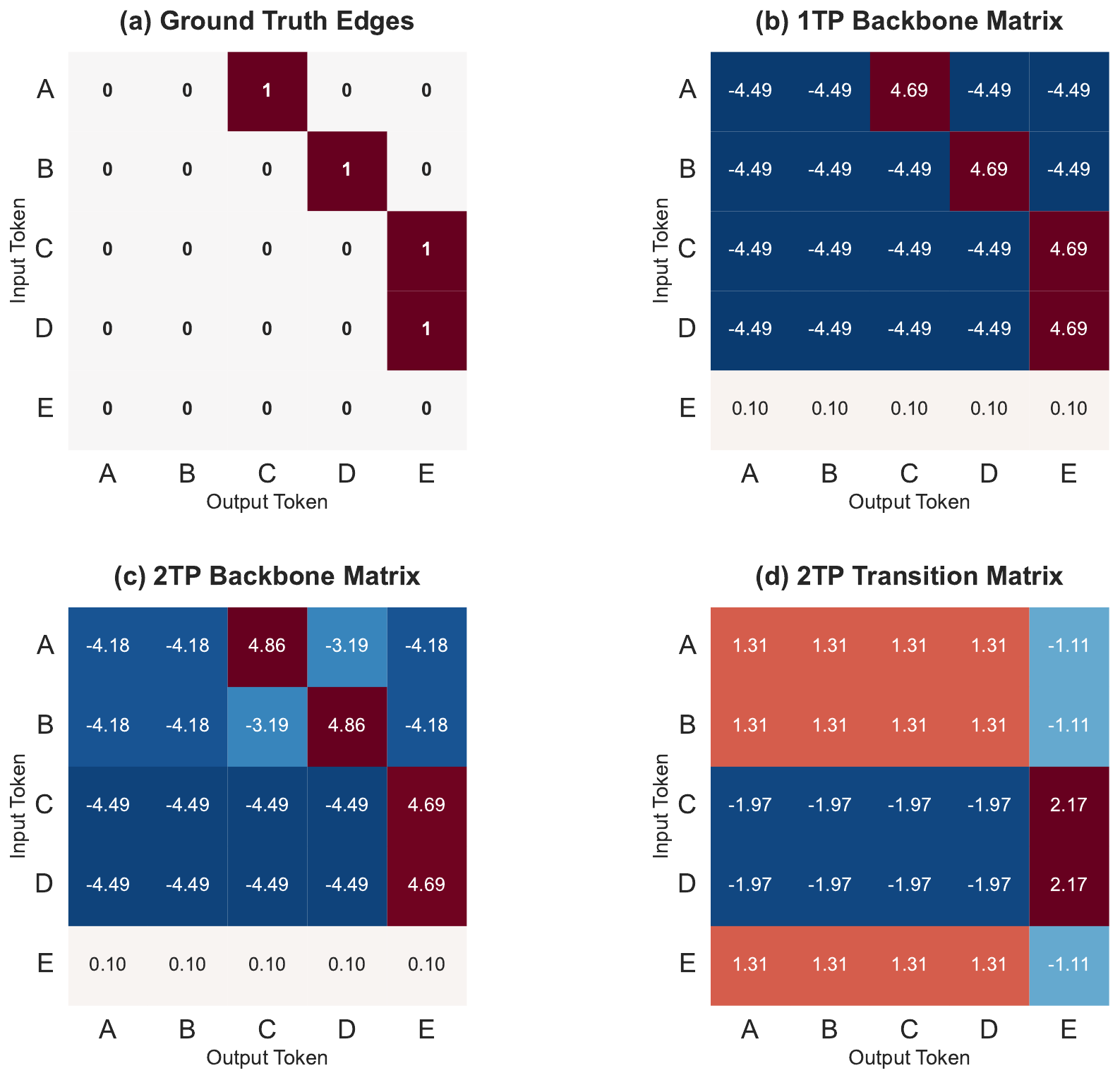}
    \caption{\textbf{Visualization of learned weights.} Under 2TP, unobserved cross-path transitions ($A \to D$, $B \to C$) are strengthened relative to 1TP.}
    \label{fig:linear_weights}
\end{figure}

As shown in Figure~\ref{fig:linear_weights}, under 1TP, $\bm{W}^B$ learns only observed transitions like $A\to C$ (Figure 3b). Under 2TP, $\bm{W}^T$ captures two-step mappings from $C,D$ to $E$ (Figure 3d), while $\bm{W}^B$ also strengthens the unobserved transition $A\to D$ (Figure 3c).

This directly illustrates Theorem~\ref{thm:mtp}: since both $C$ and $D$ lead to the shared future target $E$, the gradient for predicting $E$ backpropagates through $\bm{W}^T$ to both states, simultaneously strengthening the weights from $A$. Thus, when future targets coincide, MTP couples gradients across paths and updates transitions absent from the training data.

\subsection{Representation Alignment under Multi-Token Supervision}
\label{subsec:geometric_alignment}

We next investigate how multi-step supervision affects hidden state alignment.

Representation alignment is evaluated on two types of graphs:  
\begin{itemize}
    \item \textbf{ER (Erdős--Rényi Graphs)}: Random directed graphs capturing pure topological structure without spatial semantics. 
    \item \textbf{USG (Urban Street Graphs)}: Planar road networks with node IDs reflecting approximate geography, enabling assessment of both topological and spatial continuity~\cite{barthelemy2025universal}.
\end{itemize}

The navigation task is framed as conditional sequence generation. Given a start node $S$ and a goal node $G$, forming the context $[S, G]$, the model autoregressively predicts stepwise increments $(\text{inc}_1, \dots, \text{inc}_T)$, where each increment represents an action and node IDs are computed recursively as $u_t = u_{t-1} + \text{inc}_t$. A trajectory is valid if each increment corresponds to an existing edge $(u_{t-1}, u_t)$ and the final node reaches the goal $u_T = G$. During training, sequences $[S, G, \text{inc}_1, \dots, \text{inc}_T]$ serve as both input and autoregressive targets.

Reachable node pairs are split into 90\% training and 10\% test sets, with training paths generated via K-shortest paths, detours, and corrective strategies. On 100-node graphs, a 6-layer Transformer (6 attention heads, hidden dimension 120) is trained for 20,000 iterations, achieving $\sim$97\% accuracy. This confirms that the model sufficiently masters the task to support subsequent representation analysis.  
The code is available at~\url{https://github.com/QiminZhong/LSE-MTP}.

\begin{table*}[htbp]
\centering
\caption{\textbf{Belief compression on ER and USG graphs.} Values report hidden-state similarity for trajectories sharing the same goal $G$ and next-step position $P$ under different control conditions; $=$ denotes “same” and $\neq$ denotes “different”.}
\label{tab:exp2}
\small
\setlength{\tabcolsep}{5pt}
\begin{tabular}{ll | cccc || cccc}
\toprule
\multirow{2}{*}{\textbf{Model}} & \multirow{2}{*}{\textbf{$K$}} & \multicolumn{4}{c||}{\textbf{ER (Erdős--Rényi Graph)}} & \multicolumn{4}{c}{\textbf{USG (Urban Street Graph)}} \\
\cmidrule{3-6} \cmidrule{7-10}
& & \textbf{$G=, P=$} & \textbf{$G=, P\neq$} & \textbf{$G\neq, P=$} & Baseline & \textbf{$G=, P=$} & \textbf{$G=, P\neq$} & \textbf{$G\neq, P=$} & Baseline \\
\midrule
NTP (1TP) & 1 & 0.29 & 0.11 & 0.09 & 0.01 & 0.22 & 0.09 & 0.10 & 0.03 \\
\midrule
MTP & 2 & 0.39 & 0.23 & 0.11 & 0.05 & 0.28 & 0.10 & 0.11 & 0.02 \\
MTP & 3 & 0.43 & 0.28 & 0.14 & 0.07 & 0.30 & 0.11 & 0.10 & 0.02 \\
MTP & 4 & \textbf{0.44} & \textbf{0.30} & \textbf{0.15} & \textbf{0.08} & \textbf{0.32} & \textbf{0.12} & \textbf{0.09} & \textbf{0.03} \\
\midrule
\midrule
LSE-MTP & 2 & 0.40 & 0.25 & 0.12 & 0.05 & 0.34 & 0.13 & 0.16 & 0.05 \\
LSE-MTP & 3 & 0.44 & 0.31 & 0.13 & 0.06 & 0.37 & 0.14 & 0.17 & 0.06 \\
LSE-MTP & 4 & \textbf{0.46} & \textbf{0.34} & \textbf{0.16} & \textbf{0.09} & \textbf{0.38} & \textbf{0.15} & \textbf{0.17} & \textbf{0.06} \\
\bottomrule
\end{tabular}
\end{table*}

\subsubsection{States with the Same Future Become Aligned}
\label{sec:5.2.1}

To quantify the effect of multi-step supervision on representation alignment, 
we introduce \textbf{Structure Gain}, measuring how closely states leading to the same future are embedded in latent space. 
The metric focuses on \emph{$k$-step future equivalent} state pairs ($\mathbf{h}_1 \sim_k \mathbf{h}_2$)—corresponding to the same token at step $k$ but with different next-step targets—thus removing the confounding effect of immediate target agreement.
Structure Gain is defined as the improvement in average cosine similarity of such state pairs relative to a random baseline. 
We compare a standard next-token prediction model (NTP, 1TP) with multi-token prediction models (MTP) trained with different prediction horizons $K \in \{2, 3, 4\}$, and evaluate at $k \in \{2, 3, 4\}$.

In each experiment, we randomly sample 4{,}000 training trajectories, extract normalized hidden states from the final Transformer layer, and construct pairs satisfying $k$-step future equivalence.

Table~\ref{tab:exp1} shows that NTP exhibits low structure gain, indicating poor alignment of states sharing the same future. 
MTP models achieve substantially higher structure gain, with the effect strongest when training and evaluation horizons match ($k$). 
This trend supports Lemma~\ref{lem:mtp}: multi-token prediction induces cross-path gradient coupling, progressively converging states that lead to the same future in latent space.

\subsubsection{Path Histories Are Compressed into a Unified Belief Representation}
\label{sec:5.2.2}

Beyond aligning future-equivalent states, we further investigate whether the model compresses diverse path histories into a unified internal representation. To this end, we introduce the \textbf{Belief Compression} metric, which quantifies the similarity of hidden states corresponding to trajectories that share the same goal $G$ and, at the next step, reach the same position $P$, avoiding bias from identical immediate actions. This metric assesses whether the model can abstract away variations from different traversal histories and form a coherent internal \emph{belief state}.

In our experiments, we randomly sample 4,000 training paths and evaluate all models on the same dataset. To examine the influence of goal and positional information in the representations, we introduce three control groups:
(a) same goal, different positions ($G = , P \neq $);  
(b) different goals, same position ($G \neq , P = $);  
(c) different goals, different positions (baseline).  

Table~\ref{tab:exp2} summarizes the results. As the prediction horizon increases, MTP models exhibit higher hidden-state similarity under the same-goal, same-position condition ($G = , P = $), indicating that diverse path histories are compressed into a consistent belief representation. In contrast, the control settings and baseline show only minor increases, suggesting that compression is primarily driven by shared future outcomes.

\begin{table*}[htbp]
\centering
\caption{\textbf{Next-step probability coupling on ER and USG graphs.} ISP and Legal Prob report the probability of illegal shortcuts and valid actions for trajectories sharing a common future.}
\label{tab:exp3}
\small
\setlength{\tabcolsep}{10pt}
\begin{tabular}{ll | cc || cc}
\toprule
\multirow{2}{*}{\textbf{Model}} & \multirow{2}{*}{\textbf{$K$}} & \multicolumn{2}{c||}{\textbf{ER (Erdős--Rényi Graph)}} & \multicolumn{2}{c}{\textbf{USG (Urban Street Graph)}} \\
\cmidrule{3-4} \cmidrule{5-6}
& & ISP $\downarrow$ & Legal Prob $\uparrow$ & ISP $\downarrow$ & Legal Prob $\uparrow$ \\
\midrule
NTP (1TP) & 1 & $2.7 \times 10^{-5}$ & \textbf{0.995} & $2.2 \times 10^{-5}$ & \textbf{0.998} \\
\midrule
MTP & 2 & $4.2 \times 10^{-5}$ & 0.994 & $4.9 \times 10^{-5}$ & 0.996 \\
MTP & 3 & $7.8 \times 10^{-5}$ & 0.992 & $7.3 \times 10^{-5}$ & 0.994 \\
MTP & 4 & \textbf{1.04 $\mathbf{\times 10^{-4}}$} & 0.985 & \textbf{1.33 $\mathbf{\times 10^{-4}}$} & 0.989 \\
\midrule
\midrule
LSE-MTP & 2 & $3.0 \times 10^{-5}$ & 0.995 & $4.1 \times 10^{-5}$ & 0.997 \\
LSE-MTP & 3 & $5.1 \times 10^{-5}$ & 0.993 & $4.8 \times 10^{-5}$ & 0.996 \\
LSE-MTP & 4 & \textbf{6.3 $\mathbf{\times 10^{-5}}$} & 0.990 & \textbf{8.2 $\mathbf{\times 10^{-5}}$} & 0.994 \\
\bottomrule
\end{tabular}
\end{table*}

\subsubsection{A Pitfall: Probability Coupling in Next-Step Predictions}
\label{sec:5.2.3}

While MTP promotes representational alignment, it can introduce a teleological bias where the model prioritizes future outcomes over immediate constraints. Theorem~\ref{thm:mtp} indicates that when distinct action sequences converge on the same future action token $f$, MTP induces predictive coupling within the next-step distribution. This effect can blur the distinction between feasible increments and illegal shortcuts—action tokens that move toward $f$ but are invalid at the current state.

We evaluate this behavior using $10,000$ samples. Each test case involves a pair of action tokens $(a, a')$ that share a common future action token $f$ within two to four steps. In each pair, $a$ is a valid increment along a legal edge, while $a'$ is an illegal shortcut to an unconnected node. The model's performance is measured by \textbf{Illegal Shortcut Probability (ISP)}, the probability of the forbidden increment $a'$, and \textbf{Legal Prob}, the total probability assigned to all valid actions.

Even a single token prediction error can cause the entire sequence to fail. As shown in Table~\ref{tab:exp3}, ISP gradually increases while Legal Prob decreases as the prediction horizon grows. The ISP reported in the table counts only illegal actions pointing to a single future token $f$, but the overall decline in Legal Prob reflects the cumulative effect of probability coupling across all potential illegal actions.

This observation is consistent with our theoretical perspective: while MTP promotes alignment of trajectories sharing a common future, it also blurs distinctions among feasible next-step predictions, resulting in illegal shortcuts due to prioritizing future alignment over immediate constraints.

\paragraph{Remark.} Although the above experiments only present results from a single ER graph and a single USG graph, these phenomena are consistently observed across all generated graph instances.

\section{Why LSE-MTP}
\label{sec:why_lse_mtp}

The core motivation for LSE-MTP is to mitigate the \textit{teleological bias} inherent in standard Multi-Token Prediction (MTP) under discrete-token supervision. In standard MTP, the gradient from the cross-entropy loss $\mathcal{L}_{ce}$ is focused solely on the discrete target token $u_{n+k}$, creating a "blind spot" regarding the feasibility of the intermediate path. This often encourages the model to adopt illegal shortcuts in latent space that violate structural constraints of the environment.

LSE-MTP addresses this issue by using the future hidden state $\mathbf{h}_{n+k-1}$ as a topological anchor. Since both $\hat{\mathbf{h}}_{n,k}$ and $\mathbf{h}_{n+k-1}$ are decoded by the shared output head to predict the same future token $u_{n+k}$, they are encouraged to occupy a consistent position in latent space. Targeting $\mathbf{h}_{n+k-1}$ is advantageous because it is generated through teacher forcing, thereby incorporating the ground-truth tokens $u_{n+1:n+k-1}$ along the path. This idea draws inspiration from \citet{lamb2016professor}, where teacher-forced hidden states serve as a continuous supervisor to regularize the model's self-generated trajectories. By aligning $\hat{\mathbf{h}}_{n,k}$ to $\mathbf{h}_{n+k-1}$, the latter acts as a grounded proxy that captures the structural rules of the environment that a "jump-step" prediction might otherwise bypass. 
In practice, LSE-MTP incurs almost zero additional computational cost compared to standard MTP, as detailed in Appendix~\ref{sec:efficiency_analysis}.

This alignment mechanism also resonates with the principles of the Joint-Embedding Predictive Architecture (JEPA) \citep{lecun2022path} and Contrastive Predictive Coding (CPC) \citep{oord2018representation}, which advocate predicting future dynamics in latent space rather than in the observation space. By performing latent backpropagation, structural information from the true trajectory is directly injected into the predictive transition layers. To stabilize training, the semantic loss $\mathcal{L}_{semantic}$ acts as a complementary regularizer, anchoring predictions to the static embedding manifold. This dual-grounding mechanism mitigates hallucinations by enhancing \textbf{Belief Compression} for identical states while simultaneously reducing \textbf{Illegal Shortcut Probability (ISP)} and increasing the probability assigned to valid actions, as evidenced in Tables~\ref{tab:exp2} and~\ref{tab:exp3}. A more comprehensive sensitivity analysis of hyperparameters and the generalizability of LSE-MTP to unseen paths can be found in Appendix~\ref{app:further_analysis}.

\begin{table*}[htbp]
\centering
\caption{\textbf{Evaluation results on real-world Manhattan Taxi Ride Modeling.} Values are reported as mean (standard deviation).}
\label{tab:results}
\resizebox{\textwidth}{!}{
\begin{tabular}{lccccccc}
\toprule
\textbf{Model} & \makecell[b]{\textbf{Valid} \\ \textbf{Trajectories}} & \makecell[b]{\textbf{Current} \\ \textbf{State Probe}} & \makecell[b]{\textbf{State-wise} \\ \textbf{Similarity}} & \makecell[b]{\textbf{Compression} \\ \textbf{Precision}} & \makecell[b]{\textbf{Distinction} \\ \textbf{Precision}} & \makecell[b]{\textbf{Distinction} \\ \textbf{Recall}} & \makecell[b]{\textbf{Detour} \\ \textbf{Robustness}} \\
\midrule
\textit{Sample Size} & \textit{1000 trials} & \textit{1000 seqs} & \textit{5000 trials} & \textit{1000 trials} & \textit{1000 trials} & \textit{1000 trials} & \textit{1000 trials} \\
\midrule
1TP (baseline) & 0.993 (0.003) & 0.926 (0.001) & 0.693 (0.139) & 0.108 (0.011) & \textbf{0.357} (0.015) & 0.210 (0.010) & 0.692 (0.016) \\
4TP & 0.997 (0.002) & 0.964 (0.000) & 0.722 (0.119) & 0.119 (0.011) & 0.298 (0.014) & 0.195 (0.010) & 0.708 (0.014) \\
8TP & 0.995 (0.002) & 0.964 (0.000) & 0.820 (0.098) & 0.114 (0.011) & 0.293 (0.014) & 0.182 (0.009) & 0.716 (0.014) \\
LSE-4TP & 0.997 (0.002) & 0.943 (0.001) & 0.791 (0.127) & 0.135 (0.012) & 0.327 (0.015) & \textbf{0.213} (0.011) & 0.727 (0.014) \\
LSE-8TP & \textbf{0.998} (0.001) & \textbf{0.967} (0.000) & \textbf{0.851} (0.091) & \textbf{0.143} (0.012) & 0.285 (0.014) & 0.201 (0.010) & \textbf{0.733} (0.014) \\
\midrule
True world model & 1.000 (0.000) & 1.000 (0.000) & 1.000 (0.000) & 1.000 (0.000) & 1.000 (0.000) & 1.000 (0.000) & 1.000 (0.000) \\
\bottomrule
\end{tabular}
}
\end{table*}

\section{Evaluation on Real-World Manhattan Taxi Ride Modeling}
\label{sec:evaluation}

We evaluate our model on the Manhattan taxi trajectory benchmark introduced by \citet{vafa2024evaluating},
where city streets are abstracted as a graph with explicit topological constraints.
Given a start and a destination, models are required to generate complete routes that are graph-consistent.
This benchmark is suited for assessing the coherence of latent world models, as it reveals failures that are hard to detect with next-step prediction alone, such as infeasible paths or broken connectivity.

We train and evaluate the model on the shortest-paths dataset derived from this benchmark.
All models adopt a Transformer architecture with 12 layers, 12 attention heads, and 768-dimensional embeddings,
and are trained for 30 epochs to ensure convergence.

Most of the following metrics are adopted from \citet{vafa2024evaluating} to assess the model’s world modeling capability.
(1) \textbf{Valid Trajectories} measures the fraction of complete sequences generated on unseen start--goal pairs that satisfy all street topology constraints and successfully reach the destination.
(2) \textbf{Current State Probe} evaluates the accuracy of a linear classifier trained to predict the current node from the final-layer hidden representation.
(3) \textbf{State-wise Similarity} computes the average cosine similarity between final-layer hidden states when two different paths reach the same node with the same goal.
(4) \textbf{Compression Precision} is the fraction of continuations generated from one path that are assigned a prediction probability above a threshold ($\epsilon = 0.01$) under the other path’s context, when two paths reach the same node with the same goal.
(5) \textbf{Distinction Precision} measures, for two paths that differ in node or goal, the fraction of continuations that receive probability above $\epsilon = 0.01$ for only one path and correctly reflect the underlying map legality.
(6) \textbf{Distinction Recall} evaluates, for continuations that are legal for only one of the two paths in the true map, the proportion of cases where the model correctly assigns a probability above $\epsilon = 0.01$ to one path and below the threshold to the other.
Finally, (7) \textbf{Detour Robustness} computes the fraction of generated trajectories that remain valid and reach the goal when random non-Top-1 but legal turns are injected during generation with fixed probabilities $p = 0.01$.

Table~\ref{tab:results} presents the evaluation on real-world Manhattan taxi trajectories. Multi-Token Prediction (MTP) improves both \textbf{state-wise similarity} and \textbf{compression precision}, indicating that trajectories sharing future dynamics are mapped to more consistent latent representations. This demonstrates that MTP effectively captures shared-future structure in the latent space. However, this increased alignment comes with a slight decrease in \textbf{distinction precision}, reflecting the inherent trade-off between aligning shared-future trajectories and preserving fine-grained state differences.

Incorporating LSE as a constraint on MTP mitigates this trade-off by grounding latent states in teacher-forced future representations. LSE further enhances \textbf{compression precision} while preserving or even boosting \textbf{distinction precision}, yielding a more balanced latent space that aligns shared-future trajectories without collapsing structurally relevant distinctions. The improved \textbf{detour robustness} also indicates that the learned latent dynamics are coherent and resilient to trajectory perturbations, enabling more robust trajectory planning.

\section{Conclusion and Discussion}

In this work, we study how multi-token prediction (MTP) shapes the internal representations of sequence models for latent world modeling. Our theoretical and empirical analyses reveal a key tension: while MTP promotes convergence toward shared-future belief states, discrete token supervision can induce structural hallucinations that disrupt latent dynamics. To address this, we propose Latent Semantic Enhancement MTP (LSE-MTP), which grounds multi-step predictions in teacher-forced latent trajectories and semantic embeddings. Experiments on synthetic graphs and real-world Manhattan taxi data show that LSE-MTP improves representation alignment, belief compression, and robustness, while reducing illegal shortcuts.

These results underscore that token-level accuracy alone is insufficient for coherent world modeling. By enforcing structurally consistent latent trajectories, LSE-MTP effectively bridges discrete supervision and continuous representations. This enables models to extend their predictive horizon while better preserving the local constraints that define the environment.

These structural challenges also apply to large-scale NLP tasks. In open-ended language, environmental constraints are not explicitly defined but emerge from an implicit logical and semantic manifold that governs coherence, causality, and plausibility. By leveraging teacher-forced hidden states, LSE-MTP captures coherent semantic trajectories along this manifold, modeling stepwise dependencies and gradual contextual evolution. Anchoring multi-step latent predictions to these trajectories, LSE-MTP provides a structural alignment signal that mitigates abrupt semantic shifts, encouraging the model to better integrate intermediate contextual cues and improve long-horizon coherence.

Such latent-space regularization is particularly crucial for tasks that require precise state tracking, such as narrative understanding, code generation, or mathematical reasoning~\citep{schuster2023entity, li2025how}. These tasks demand that models maintain consistent representations of entities, variables, or arguments over extended contexts. By regularizing latent trajectories, LSE-MTP helps transform language models from local pattern matchers into coherent internal simulators capable of reliable long-horizon reasoning.

\newpage
\section{Limitations}

First, our experimental evaluation is primarily focused on structured graph navigation and path-planning tasks, with its applicability to open-ended natural language problems with higher levels of abstraction and more complex semantic dynamics not yet fully explored. Second, we have only analyzed and experimented with the widely used MTP model, without conducting a systematic comparison with other models and methods aimed at enhancing latent representation consistency, such as reinforcement learning objectives, contrastive representation learning, or explicit state-space modeling. Finally, our theoretical perspective relies on a linearized gradient flow approximation, which, while capturing the core trends of the training dynamics, may not fully reflect the complex nonlinear behavior of large-scale Transformer models.

\section*{Acknowledgments}
This work is supported by thet National Natural
Science Foundation of China (Grant No.
62276171, 62476173, 62532007), Guangdong Basic
and Applied Basic Research Foundation (Grant
No. 2024A1515011938 and 2020B1515120028), Shenzhen Fundamental Research Project (Grant No. ZDCY20250901110940006, JCYJ20240813\-141503005,  JCYJ20240813142610014) Major Special
Project for Philosophy and Social Sciences Research
of the Ministry of Education (Grant No. 2025JZDZ010). CCF-Huawei Populus Grove Fund (Grant No. CCF-HuaweiFM2024004).

\bibliography{main}

\clearpage
\appendix

\begin{center}
    \Large\bfseries Appendix
\end{center}

\section{Related Works}
\label{app:related-works}

\subsection{World Models in Language Modeling}

The debate over whether Large Language Models (LLMs) are "stochastic parrots" \citep{bender2021danger} or possess emergent \textit{world models} remains central to NLP \citep{li2023emergent, patel2022mapping}. Probing studies suggest that neural language models can indeed develop implicit representations of meaning and world states even when trained solely on text \citep{li2021implicit}. While Transformers can internalize structural invariants like game states \citep{li2023emergent} or geographical coordinates \citep{gurnee2023language}, and even demonstrate a nascent grasp of fundamental physical concepts \citep{tang2024can}, they are often reactive rather than truly predictive. Recent studies highlight failures in multi-step causal reasoning and state tracking \citep{valmeekam2023planbench, wu2024can}, showing fragility under structural perturbations \citep{vafa2024evaluating}. Beyond simple internalization, discovering structured and modular world models from data is a prerequisite for reliable planning. This shift toward causal modularity facilitates the emergence of consistent belief states \citep{lei2022variational}, encouraging paradigms that frame reasoning as an explicit planning process over an internal world model \citep{hao2023reasoning}, and motivating architectures that move from local co-occurrence statistics toward explicit environment dynamics \citep{hafner2025mastering, lecun2022path}.

\subsection{Multi-Token Prediction}

Multi-Token Prediction (MTP) improves upon standard next-token prediction by supervising multiple future tokens, which enhances performance on reasoning benchmarks \citep{gloeckle2024better}. By incentivizing the model to anticipate future sequence fragments during pre-training, MTP builds on the intuition that future n-gram prediction can foster more robust contextual representations and sequence-level planning \citep{qi2020prophetnet}. Theoretically, MTP fosters longer-range dependencies and "look-ahead" foresight \citep{olsson2022induction, cai2024medusa}. In graph-based planning, MTP prevents local shortcuts via multi-hop trajectory mixtures \citep{arvid2025language} and captures transitive reachability by encoding multi-step adjacency \citep{zhong2024understanding}. This approach is rooted in earlier sequence-level optimization techniques like MIXER \citep{ranzato2016sequence}, which aim to mitigate exposure bias \citep{bachmann2024the} by narrowing the distributional gap between teacher-forcing training and autoregressive inference \citep{zhang2019bridging}, preventing the accumulation of errors during rollout. However, behavioral gains do not guarantee latent consistency, as models may still learn shortcuts that bypass underlying environmental rules \citep{geirhos2020shortcut}. Our work explores this risk of "structural hallucinations" and proposes latent grounding as a necessary stabilizer.

\subsection{Latent Consistency and State-Space Alignment}

Reliable world modeling requires internal representations to evolve consistently with environmental dynamics. In autonomous intelligence, architectures such as JEPA \citep{lecun2022path} and Dreamer \citep{hafner2025mastering} advocate predicting in latent space rather than observation space, thereby filtering out task-irrelevant noise. NextLat \citep{teoh2025nextlatent} extends this principle to Transformers through self-supervised latent-state prediction, encouraging models to learn compressed belief states and form compact internal world models. Recent efforts have further explored aligning hidden states with symbolic world structures \citep{zhu2024language, garrido2024learning} or leveraging tokenized latent states \citep{zhai2025recurrent}. Similarly, methods such as Semformer \citep{yin2024semformer} promote semantic planning by predicting future latent representations generated by an auxiliary autoencoder. Building on these insights, LSE-MTP draws inspiration from knowledge distillation \citep{hinton2015distilling} and consistency models \citep{song2023consistency} to improve transition consistency by anchoring predictions to ground-truth hidden states.

\subsection{Graph-based Reasoning and Navigation}

Graphs provide a rigorous testbed for world models due to their explicit transition rules \citep{li2023emergent, wu2024can}. Navigating these environments requires models to maintain logical consistency through structure-aware architectures that can handle complex relational constraints and densification \citep{lin2024salmon}. Recent benchmarks, such as Manhattan taxi trajectories \citep{vafa2024evaluating}, require models to adhere to real-world topology over long horizons. Despite generating fluent paths, LLMs often prioritize statistical patterns over topological constraints, leading to planning failures during detours \citep{valmeekam2023planbench, stechly2025on}. Such failures emphasize the need for graph-based verification mechanisms that can audit reasoning chains against the underlying connectivity to ensure path validity \citep{chen2024graphreason}. While specialized fine-tuning or prompting techniques like structuring internal evidence into a graph of thoughts can improve performance \citep{fatemi2024talk, wen2024mindmap}, the fundamental challenge of latent state legality remains. We utilize synthetic and real-world graphs to demonstrate how latent grounding prevents models from taking illegal shortcuts that violate connectivity.

\section{Derivations and Proofs}
\label{app:proofs}

This appendix rigorously characterizes the gradient dynamics, detailing assumptions, validity conditions, and proofs for the established theorems.

\subsection{Validity of Linearized Analysis}
\label{app:validity}
To understand how the training process shapes internal representations, we analyze the model's behavior through the lens of the linearized regime, also known as \textit{lazy training}~\citep{chizat2019lazy}. Deep neural networks, like Transformers, are notoriously complex and non-linear, making their training dynamics difficult to track mathematically. However, a key theoretical insight in deep learning is that as a network becomes sufficiently wide, its individual weights $\theta$ only need to change by a tiny amount from their initial values $\theta_0$ to significantly reduce the training loss. In this "lazy" state, we can accurately approximate the network’s output, specifically the hidden state $f_\theta(H)$, using a first-order Taylor expansion:
\begin{equation}
f_\theta(H) \approx f_{\theta_0}(H) + \nabla_\theta f_{\theta_0}(H)^\top (\theta - \theta_0).
\end{equation}

This approximation effectively treats the complex network as a linear model during the early stages of training. The primary advantage of this approach is that it allows us to define the \textbf{Neural Tangent Kernel (NTK)}, a mathematical object that remains approximately constant during training. The NTK acts like a \textit{geometric map} of the representation space, determining how an update on one input, such as a specific history $H_i$, influences the representation of another, $H_j$. By assuming this kernel is stable, we can derive closed-form proofs for how gradients flow through the model.

While real-world, finite-width Transformers eventually move beyond this linear phase to perform \textit{feature learning}, the linearized analysis remains a powerful tool for our purposes. It provides a clear, qualitative explanation of the instantaneous directional pressure, which represents the immediate "force" that the Multi-Token Prediction (MTP) objective exerts on hidden states. By capturing the direction in which the loss function pushes representations at any given moment, this framework reveals the mathematical root of the gradient coupling and representational contraction observed in our empirical experiments.

\subsection{Evolution Dynamics and Notation}
\label{app:evolution_dynamics}
To track how hidden states $\mathbf{h}$ evolve during training, we study their dynamics under \textbf{gradient flow}. Let $\mathbf{h} = f_\theta(H) \in \mathbb{R}^d$ denote the hidden representation of a history $H$. The continuous-time optimization of parameters is controlled by a learning rate $\eta > 0$, and the weight evolution is given by:
\begin{equation}
\dot{\theta} = \frac{d\theta}{dt} = -\eta \nabla_\theta \mathcal{L},
\end{equation}
where $\mathcal{L}$ is the loss function. Applying the chain rule, the velocity of the hidden state (the rate of change over time) satisfies:
\begin{equation}
\dot{\mathbf{h}} 
= \nabla_\theta f_\theta(H) \dot{\theta}
= -\eta \nabla_\theta f_\theta(H) \nabla_\theta \mathcal{L}.
\end{equation}
Since the loss depends on the weights $\theta$ primarily through the representation $\mathbf{h}$, we can further decompose the weight gradient using the chain rule again:
\begin{equation}
\nabla_\theta \mathcal{L} = \nabla_\theta f_\theta(H)^\top \nabla_{\mathbf{h}} \mathcal{L}.
\end{equation}
Substituting this back into the velocity equation yields:
\begin{equation}
\begin{aligned}
\dot{\mathbf{h}}
&= -\eta \left[\nabla_\theta f_\theta(H)\nabla_\theta f_\theta(H)^\top\right] \nabla_{\mathbf{h}}\mathcal{L} \\
&= -\eta\,\mathbf{K}(\mathbf{h},\mathbf{h})\nabla_{\mathbf{h}}\mathcal{L},
\end{aligned}
\end{equation}
where
\begin{equation}
\mathbf{K}(\mathbf{h}_i, \mathbf{h}_j) = \nabla_\theta f_\theta(H_i) \nabla_\theta f_\theta(H_j)^\top \in \mathbb{R}^{d \times d}
\end{equation}
is the NTK matrix block, which measures the geometric correlation between the gradients of two different history samples $H_i$ and $H_j$. 

Intuitively, this expression shows that hidden-state updates are driven by the loss gradient $\nabla_{\mathbf{h}}\mathcal{L}$ and modulated by the kernel $\mathbf{K}$, which captures geometric correlations between different histories. When $\mathbf{K}$ exhibits strong cross-history coupling, the corresponding representations are forced to evolve jointly, providing the core mechanism behind the representational contraction.

\subsection{Proof of Theorem 1}
\label{app:proof_thm_ntp}

\textbf{Theorem 1}
\textit{
Under the NTP loss $\mathcal{L}_{\text{NTP}}$, the contractive condition $\dot{\mathcal{D}} \le 0$ holds primarily for $1$-step equivalent states ($\mathbf{h}_1 \sim_1 \mathbf{h}_2$). For states with different next-step targets, the gradients $\nabla_{\mathbf{h}} \mathcal{L}$ tend to point in opposite directions, preserving representational separation.
}

\paragraph{Proof.}
To analyze the convergence of hidden states, we define the representational distance as $\mathcal{D} = \|\mathbf{h}_1 - \mathbf{h}_2\|^2$. The time derivative of this distance, which represents the rate at which states move toward or away from each other, is calculated as:
\begin{equation}
    \dot{\mathcal{D}} = \frac{d}{dt} \|\mathbf{h}_1 - \mathbf{h}_2\|^2 = 2(\mathbf{h}_1 - \mathbf{h}_2)^\top (\dot{\mathbf{h}}_1 - \dot{\mathbf{h}}_2).
\end{equation}

By substituting the hidden-state velocity formula $\dot{\mathbf{h}} = -\eta \mathbf{K} \nabla_{\mathbf{h}} \mathcal{L}$ derived in Section \ref{app:evolution_dynamics}, the dynamics are expressed as:
\begin{equation}
\begin{split}
\dot{\mathcal{D}} = -2\eta (\mathbf{h}_1 - \mathbf{h}_2)^\top 
\big[ 
& \mathbf{K}(\mathbf{h}_1, \mathbf{h}_1) \nabla_{\mathbf{h}_1} \mathcal{L} \\
& - \mathbf{K}(\mathbf{h}_2, \mathbf{h}_2) \nabla_{\mathbf{h}_2} \mathcal{L} 
\big],
\end{split}
\end{equation}
where $\eta$ denotes the learning rate, $\mathcal{L}$ is the loss function, and $\mathbf{K}(\mathbf{h}_1, \mathbf{h}_1), \mathbf{K}(\mathbf{h}_2, \mathbf{h}_2)$ are the auto-kernel blocks for histories $H_1, H_2$.

\paragraph{Assumption 1 (Local Kernel Smoothness).} For nearby states, we assume the kernel varies smoothly such that $\mathbf{K}(\mathbf{h}_1, \mathbf{h}_1) \approx \mathbf{K}(\mathbf{h}_2, \mathbf{h}_2) \approx \mathbf{K}$, where $\mathbf{K}$ is a positive semi-definite matrix. This assumption implies that the geometric properties of the representation space are locally stable, ensuring consistent sensitivity to parameter updates for nearby histories.

Using this assumption and letting $\Delta\mathbf{h} = \mathbf{h}_1 - \mathbf{h}_2$, the dynamics of the representational distance simplify to:
\begin{equation}
    \dot{\mathcal{D}} \approx -2\eta \Delta\mathbf{h}^\top \mathbf{K} \left( \nabla_{\mathbf{h}_1} \mathcal{L} - \nabla_{\mathbf{h}_2} \mathcal{L} \right) \label{eq:distance_dynamics}.
\end{equation}

To further simplify the gradient difference term, we consider the gradient $\nabla_{\mathbf{h}} \mathcal{L}$ as a vector-valued function of $\mathbf{h}$. Since $\mathbf{h}_1$ and $\mathbf{h}_2$ are assumed to be in close proximity, we can apply a first-order Taylor expansion to the gradient $\nabla_{\mathbf{h}_1} \mathcal{L}$ around the point $\mathbf{h}_2$:
\begin{equation}
    \nabla_{\mathbf{h}_1} \mathcal{L} \approx \nabla_{\mathbf{h}_2} \mathcal{L} + \left[ \nabla_{\mathbf{h}}^2 \mathcal{L} \right] (\mathbf{h}_1 - \mathbf{h}_2),
\end{equation}
where $\nabla_{\mathbf{h}}^2 \mathcal{L}$ is the Hessian matrix of the loss function, denoted as $\mathbf{H}_{\text{loss}}$. This matrix captures the local curvature of the optimization landscape.

By rearranging the above expansion, we obtain an approximation for the gradient difference:
\begin{equation}
    \nabla_{\mathbf{h}_1} \mathcal{L} - \nabla_{\mathbf{h}_2} \mathcal{L} \approx \mathbf{H}_{\text{loss}} \Delta\mathbf{h}.
\end{equation}

Finally, substituting this approximation back into Eq.~(\ref{eq:distance_dynamics}) yields the final quadratic form:
\begin{equation}
    \dot{\mathcal{D}} \approx -2\eta \Delta\mathbf{h}^\top (\mathbf{K} \mathbf{H}_{\text{loss}}) \Delta\mathbf{h}.
\end{equation}

\paragraph{Conclusion for NTP.} In Next-Token Prediction, the contractive condition $\dot{\mathcal{D}} \leq 0$ requires the gradients to converge toward a shared optimum. When target tokens differ ($u_{n+1} \neq u_{m+1}$), the gradients $\nabla_{\mathbf{h}_1} \mathcal{L}$ and $\nabla_{\mathbf{h}_2} \mathcal{L}$ point in opposite directions. As a result, $\Delta\mathbf{h}^\top (\nabla_{\mathbf{h}_1} \mathcal{L} - \nabla_{\mathbf{h}_2} \mathcal{L})$ becomes negative, leading to $\dot{\mathcal{D}} > 0$. Therefore, representations diverge unless they share the same target.

\subsection{Proof of Theorem 2}
\textbf{Theorem 2}
\textit{
Under the MTP loss $\mathcal{L}_{\text{MTP}}$, consider $k$-step future-equivalent states $\mathbf{h}_1 \sim_k \mathbf{h}_2$ with different immediate targets $u_{n+1} \neq u_{m+1}$. A $k$-step update on $\mathbf{h}_1$ induces a positive cross-update on the corresponding logit of $\mathbf{h}_2$, $\dot{z}_{y_1}(\mathbf{h}_2) > 0$, where the gradients $\nabla_{\mathbf{h}_1} \mathcal{L}^{(k)}_1$ and $\nabla_{\mathbf{h}_2} \mathcal{L}^{(k)}_1$ align through the cross-history NTK $\mathbf{K}(\mathbf{h}_1, \mathbf{h}_2)$, facilitating a predictive coupling that can partially blur the representational separation between distinct trajectories.
}

\paragraph{Proof.}
To analyze predictive coupling in MTP, we track how the $k$-step loss $\mathcal{L}^{(k)}_1$ on history $H_1$ affects the logit $z_{y_1}(\mathbf{h}_2)$ for $H_1$’s first future token $y_1$ in history $H_2$.

Under gradient flow, the parameter dynamics $\dot{\theta} = -\eta \nabla_{\theta} \mathcal{L}^{(k)}_1$ govern the evolution of $z_{y_1}(\mathbf{h}_2)$:
\begin{equation}
\begin{aligned}
\frac{d z_{y_1}(\mathbf{h}_2)}{dt} 
&= \langle \nabla_{\theta} z_{y_1}(\mathbf{h}_2), \dot{\theta} \rangle \\
&= -\eta \langle \nabla_{\theta} z_{y_1}(\mathbf{h}_2), \nabla_{\theta} \mathcal{L}^{(k)}_1 \rangle.
\end{aligned}
\label{eq:time_derivative_y1}
\end{equation}

Both gradients depend on $\theta$ only through the hidden representations $\mathbf{h}_2 = f_\theta(H_2)$ and $\mathbf{h}_1 = f_\theta(H_1)$:
\begin{equation}
\begin{aligned}
\nabla_\theta z_{y_1}(\mathbf{h}_2)
&= \nabla_{\mathbf{h}_2} z_{y_1} \, \nabla_\theta f_\theta(H_2), \\
\nabla_\theta \mathcal{L}^{(k)}_1
&= \nabla_\theta f_\theta(H_1)^\top \nabla_{\mathbf{h}_1} \mathcal{L}^{(k)}_1.
\end{aligned}
\end{equation}

Substituting into Eq.~(\ref{eq:time_derivative_y1}) gives:
\begin{equation}
\begin{aligned}
\frac{d z_{y_1}(\mathbf{h}_2)}{dt}
&= -\eta (\nabla_{\mathbf{h}_2} z_{y_1})^\top 
\mathbf{K}(\mathbf{h}_2, \mathbf{h}_1)
\nabla_{\mathbf{h}_1} \mathcal{L}^{(k)}_1,
\end{aligned}
\label{eq:ntk_update}
\end{equation}
where
\begin{equation}
\mathbf{K}(\mathbf{h}_2, \mathbf{h}_1)
= \nabla_{\theta} f_{\theta}(H_2) \nabla_{\theta} f_{\theta}(H_1)^\top
\end{equation}
is the cross-history NTK block capturing geometric coupling between the hidden representations $\mathbf{h}_2$ and $\mathbf{h}_1$.

\paragraph{Assumption 2 (Structural Alignment).}
We assume that the transition layer Jacobian $J_k$ preserves gradient orientation over $k$ steps. 
Under this assumption, the $k$-step gradient from history $H_1$, $\nabla_{\mathbf{h}_1} \mathcal{L}^{(k)}_1$, lies in the subspace spanned by the cross-history NTK $\mathbf{K}(\mathbf{h}_2, \mathbf{h}_1)$ and the $k$-step gradient of $H_2$, $\nabla_{\mathbf{h}_2} \mathcal{L}^{(k)}_2$. 
This ensures predictable interactions between gradients from different histories in multi-token prediction.

\paragraph{Conclusion for MTP.}  
The update direction is determined by the inner product in Eq.~(\ref{eq:time_derivative_y1}).  
Under the Structural Alignment Assumption, the gradient propagated from $H_1$ affects $H_2$ predictably.  
When $H_1$ and $H_2$ share the same $k$-step future token sequence, we can approximate
\begin{equation}
    \nabla_{\mathbf{h}_2} z_{y_1} \approx -\alpha \, \nabla_{\mathbf{h}_2} \mathcal{L}^{(k)}_1, \quad \alpha > 0,
\end{equation}
where the negative sign reflects that gradient descent on the loss \(\mathcal{L}^{(k)}_1\) decreases the loss but increases the corresponding logits.

Substituting this into Eq.~(\ref{eq:ntk_update}) then yields a positive cross-update:
\begin{equation}
\begin{aligned}
\frac{d z_{y_1}(\mathbf{h}_2)}{dt} 
&= -\eta (\nabla_{\mathbf{h}_2} z_{y_1})^\top \mathbf{K}(\mathbf{h}_2, \mathbf{h}_1) \nabla_{\mathbf{h}_1} \mathcal{L}^{(k)}_1 \\
&\approx -\eta \, (-\alpha \, \nabla_{\mathbf{h}_2} \mathcal{L}^{(k)}_1)^\top \mathbf{K}(\mathbf{h}_2, \mathbf{h}_1) \nabla_{\mathbf{h}_1} \mathcal{L}^{(k)}_1 \\
&= \eta \, \alpha \, (\nabla_{\mathbf{h}_2} \mathcal{L}^{(k)}_1)^\top \mathbf{K}(\mathbf{h}_2, \mathbf{h}_1) \nabla_{\mathbf{h}_1} \mathcal{L}^{(k)}_1 \\
&= \eta \, \alpha \, (\nabla_{\mathbf{h}_1} \mathcal{L}^{(k)}_1)^\top \mathbf{K}(\mathbf{h}_1, \mathbf{h}_2) \nabla_{\mathbf{h}_2} \mathcal{L}^{(k)}_1 > 0.
\end{aligned}
\end{equation}

Consequently, supervising $H_1$ via its $k$-step loss increases the probability that $H_2$ predicts the same sequence of future tokens, including $y_1$, even if $H_2$’s own next-token target $y_2$ differs.

\subsection{Proof of Lemma 1}
\label{app:proof_lem_mtp}

\textbf{Lemma 1}
\textit{
For a pair of $k$-step future-equivalent states ($\mathbf{h}_1 \sim_k \mathbf{h}_2$), 
a full-rank transition Jacobian ensures that MTP induces a stable contractive force 
with $\dot{\mathcal{D}} \le 0$. 
The resulting geometric flow is governed by $\mathbf{K}\mathbf{S}$, 
where $\mathbf{K}$ is the NTK and $\mathbf{S}$ the pull-back Hessian. 
Although $\mathbf{K}\mathbf{S}$ is generally non-symmetric, it is similar to a symmetric PSD matrix, 
implying real, non-negative eigenvalues and thus local convergence to a unified belief state.
}

\paragraph{Proof.}
We examine the contractivity of the representational distance 
\(\mathcal{D} = \|\mathbf{h}_1 - \mathbf{h}_2\|^2\) 
for histories that share a common \(k\)-step future. 
Let \(\Delta \mathbf{h} = \mathbf{h}_1 - \mathbf{h}_2\) denote the difference between hidden representations.

Under gradient flow, the evolution of each hidden state is given by
\begin{equation} \label{eq:gradient-flow}
    \dot{\mathbf{h}} = -\eta \, \mathbf{K} \nabla_{\mathbf{h}} \mathcal{L}.
\end{equation}
where \(\mathbf{K}\) is the NTK capturing the sensitivity of hidden states to parameter updates. 

The rate of change of the representational distance is obtained by differentiating 
\(\mathcal{D} = \Delta \mathbf{h}^\top \Delta \mathbf{h}\) with respect to time:
\begin{equation} \label{eq:distance_derivative}
\begin{aligned}
    \dot{\mathcal{D}} 
    &= \frac{d}{dt} (\Delta \mathbf{h}^\top \Delta \mathbf{h}) \\
    &= 2 \, \Delta \mathbf{h}^\top \frac{d}{dt}(\Delta \mathbf{h}) \\
    &= 2 \, \Delta \mathbf{h}^\top (\dot{\mathbf{h}}_1 - \dot{\mathbf{h}}_2),
\end{aligned}
\end{equation}
where we used \(\frac{d}{dt}(\mathbf{h}_1 - \mathbf{h}_2) = \dot{\mathbf{h}}_1 - \dot{\mathbf{h}}_2\).

Hence, the rate of change of the distance is determined by the projection of the hidden-state velocity difference onto the current difference \(\Delta \mathbf{h}\), providing a direct measure of convergence or divergence between the two representations.

For multi-step prediction, the loss depends on the hidden states through 
$k$ transition layers and a shared prediction head. Let 
$J_k = \frac{\partial \mathbf{z}}{\partial \mathbf{h}}$ 
denote the Jacobian from the hidden state $\mathbf{h}$ to the output logits $\mathbf{z}$, 
and 
$\mathbf{H}_{\mathrm{head}} = \frac{\partial^2 \mathcal{L}}{\partial \mathbf{z}^2}$ 
be the Hessian of the prediction head. 

Consider two hidden states \(\mathbf{h}_1\) and \(\mathbf{h}_2\) that are close in representation space. 
A first-order Taylor expansion of the gradient at \(\mathbf{h}_1\) around \(\mathbf{h}_2\) gives
\begin{equation}
    \nabla_{\mathbf{h}_1} \mathcal{L} \approx \nabla_{\mathbf{h}_2} \mathcal{L} 
    + \nabla_{\mathbf{h}}^2 \mathcal{L} \big|_{\mathbf{h}_2} \, (\mathbf{h}_1 - \mathbf{h}_2),
\end{equation}
where \(\nabla_{\mathbf{h}}^2 \mathcal{L}\) is the Hessian of the loss with respect to the hidden state. 
For multi-step prediction, the loss gradient is backpropagated through \(k\) transition layers, 
so the effective Hessian with respect to the original hidden state can be expressed via the chain rule as
\begin{equation}
    \mathbf{S} = J_k^\top \mathbf{H}_{\mathrm{head}} J_k,
\end{equation}
where \(\mathbf{H}_{\mathrm{head}} = \nabla_{\mathbf{z}}^2 \mathcal{L}_{\mathrm{head}}\) is the Hessian of the prediction head, 
and \(J_k\) maps perturbations in \(\mathbf{h}\) to the output logits \(z\) through the \((k-1)\)-th transition layer. 
Hence, the gradient difference between the two hidden states can be approximated as
\begin{equation} \label{eq:grad-diff-approx}
    \nabla_{\mathbf{h}_1} \mathcal{L} - \nabla_{\mathbf{h}_2} \mathcal{L} \approx 
    \mathbf{S} \, (\mathbf{h}_1 - \mathbf{h}_2) = \mathbf{S} \, \Delta \mathbf{h}.
\end{equation}

Intuitively, the pull-back Hessian \(\mathbf{S}\) captures how local variations in the hidden state propagate through the transition layers and the prediction head to affect the loss, effectively defining a local metric for representational contraction.

Substituting the gradient approximation Eq.~(\ref{eq:grad-diff-approx}) 
into the distance dynamics Eq.~(\ref{eq:distance_derivative}) 
and using the gradient flow Eq.~(\ref{eq:gradient-flow}), we obtain
\begin{equation} \label{eq:contractive-form}
    \dot{\mathcal{D}} \approx -2 \eta \, \Delta \mathbf{h}^\top (\mathbf{K} \mathbf{S}) \, \Delta \mathbf{h}.
\end{equation}

In this expression, \(\mathbf{K}\) quantifies how changes in model parameters affect the hidden states, 
while \(\mathbf{S}\) encodes how small variations in the hidden representations propagate through the transition layers 
and the prediction head to influence the loss. 
The product \(\mathbf{K}\mathbf{S}\) therefore defines a local metric that determines the rate and direction of contraction: 
the negative sign ensures that the component of the hidden-state difference \(\Delta \mathbf{h}\) 
along sensitive directions is reduced over time. 
As a result, representations of histories that share a common future are drawn toward each other, 
forming a stable, contractive manifold in representation space.

A key theoretical concern is that, although both the kernel $\mathbf{K}$ 
and the pull-back Hessian $\mathbf{S}$ are symmetric positive semi-definite (PSD), 
their product $\mathbf{K}\mathbf{S}$ is not guaranteed to be symmetric or PSD. 
To ensure that representations converge (i.e., $\dot{\mathcal{D}} \le 0$), 
we must verify that all eigenvalues of $\mathbf{K}\mathbf{S}$ are real and non-negative.

Assuming a locally strictly convex prediction head and a full-rank transition Jacobian, 
we have $\mathbf{S} \succ 0$. 
We can then perform a similarity transformation on $\mathbf{K}\mathbf{S}$ using the square root of the Hessian:
\begin{equation}
    \mathbf{S}^{1/2} (\mathbf{K}\mathbf{S}) \mathbf{S}^{-1/2} 
    = \mathbf{S}^{1/2} \mathbf{K} \mathbf{S}^{1/2} \triangleq \tilde{\mathbf{K}}.
\end{equation}
Since $\mathbf{K}$ is PSD and $\mathbf{S}^{1/2}$ is symmetric, 
$\tilde{\mathbf{K}}$ is symmetric and PSD. 
Because similar matrices share the same eigenvalues, we have
\begin{equation}
    \lambda_i(\mathbf{K}\mathbf{S}) = \lambda_i(\tilde{\mathbf{K}}) \ge 0.
\end{equation}

This guarantees that the dynamical system has no unstable or divergent modes. 
The induced flow generates a stable contractive force in the metric defined by $\mathbf{S}$, 
attracting representations of histories sharing a common future toward each other. 
Hence, MTP establishes a stable manifold that formally proves representational contraction.

\subsection{Boundary Conditions}
The theoretical validity of the representational contraction depends on the numerical stability of the transition Jacobian $J_k$. If $J_k$ were to suffer from rank-deficiency, the pull-back Hessian $\mathbf{S}$ would become singular, effectively halting the convergence of belief states. While this is a common failure mode in deep linear networks~\cite{saxe2013exact, pennington2017nonlinear}, modern Transformer architectures incorporate design elements that mitigate this risk. 

Residual connections and LayerNorm collectively maintain a non-zero minimum singular value~\cite{xiong2020layer}, $\sigma_{\min}(J_k) > 0$, across the hidden layers. These architectural features ensure that $\mathbf{S}$ remains positive-definite, thereby preserving the gradient flow required for the emergence of consistent internal belief states. This indicates that our theoretical findings are well-supported by the structural properties of Transformer-based world models.

\section{Detailed Dataset Construction and Experimental Setup}
\label{app:data_construction}

This section describes the procedures for generating the graph environments and the trajectory datasets used in our experiments. 

\subsection{Graph Topology Construction}
Two types of graphs are constructed: Erdős-Rényi (ER) random graphs and Planar Road Layout (USG) networks.

\begin{algorithm}[h]
\caption{ER-Random Graph Generation}
\label{alg:er_gen}
\begin{algorithmic}[1]
\Require Nodes $n$, probability $p$, boolean $is\_dag$
\Ensure Directed Graph $G = (V, E)$
\State $V \gets \{0, \dots, n-1\}$, $E \gets \emptyset$
\If{$is\_dag$}
    \State $\pi \gets \text{Random permutation of } V$ 
    \State $pos[v] \gets \text{index of } v \text{ in } \pi, \forall v \in V$
    \For{each pair $(u, v)$ where $u \neq v$}
        \If{$pos[u] < pos[v]$ \textbf{and} $\text{rand}(0, 1) < p$}
            \State $E \gets E \cup \{(u, v)\}$
        \EndIf
    \EndFor
\Else
    \For{each pair $(u, v)$ where $u \neq v$}
        \If{$\text{rand}(0, 1) < p$}
            \State $E \gets E \cup \{(u, v)\}$
        \EndIf
    \EndFor
\EndIf
\Return $G(V, E)$
\end{algorithmic}
\end{algorithm}

\paragraph{ER-Random Graphs.} We generate directed graphs using $n=100$ nodes and an edge probability $p=0.04$. For directed acyclic graphs (DAGs), edges are permitted only according to a random topological ordering. The procedure is shown in Algorithm~\ref{alg:er_gen}.

\begin{algorithm}[h]
\caption{USG-Urban Street Graph Generation}
\label{alg:US_gen}
\begin{algorithmic}[1]
\Require Nodes $n$, density $\rho$
\Ensure Directed Graph $G_{final}$
\State $Pos \gets \{ (x_i, y_i) \mid x_i, y_i \sim \mathcal{U}(0, 1) \}_{i=0}^{n-1}$
\State $G_{base} \gets \text{DelaunayTriangulation}(Pos)$
\State $T_{mst} \gets$ MinimumSpanningTree $(G_{base})$ \Comment{Ensure connectivity}
\State $E_{extra} \gets G_{base}.edges \setminus T_{mst}.edges$
\State $E_{add} \gets \text{Sample } \lfloor |E_{extra}| \cdot \rho \rfloor$ edges from $E_{extra}$
\State $G_{sparse} \gets (V, T_{mst}.edges \cup E_{add})$
\State $V_{sorted} \gets \text{Sort } V \text{ by } (x, y) \text{ lexicographically}$
\State $\phi(v) \gets \text{index of } v \text{ in } V_{sorted}$
\State $G_{relabeled} \gets \text{Relabel } G_{sparse} \text{ via mapping } \phi$
\State $G_{final} \gets \text{Convert to Directed Graph}$
\For{each edge $\{u, v\} \in G_{relabeled}$}
    \State Add $(u, v)$ and $(v, u)$ to $G_{final}$
\EndFor
\Return $G_{final}$
\end{algorithmic}
\end{algorithm}

\paragraph{USG-Urban Street Graphs.} These graphs are built using geometric triangulation and spatial relabeling. We set $n=100$ and mesh density $\rho=0.3$. The procedure is detailed in Algorithm~\ref{alg:US_gen}.

\begin{algorithm}[h]
\caption{Diverse Path Generation}
\label{alg:data_gen}
\begin{algorithmic}[1]
\Require Graph $G$, pairs $\mathcal{P}_{train}$, $K, p_{detour}, p_{rec}$
\Ensure Training Dataset $\mathcal{D}$
\State $\mathcal{D} \gets \emptyset$
\For{each pair $(s, g) \in \mathcal{P}_{train}$}
    \State $p^* \gets \text{shortest path from } s \text{ to } g$
    \State $\mathcal{P}_{K} \gets \text{top } K \text{ shortest paths from } s \text{ to } g$
    \State $\mathcal{D} \gets \mathcal{D} \cup \{[s, g, p] \mid p \in \mathcal{P}_{K}\}$
    \If{$\text{rand}() < p_{detour}$ \textbf{and} $\text{length}(p^*) \ge 4$}
        \State $v_{obs} \gets \text{random node in } p^* \setminus \{s, g\}$
        \State $G' \gets G \setminus \{v_{obs}\}$
        \State $p_{det} \gets \text{shortest path from } s \text{ to } g \text{ in } G'$
        \State $\mathcal{D} \gets \mathcal{D} \cup \{[s, g, p_{det}]\}$ \Comment{if path exists}
    \EndIf
    \If{$\text{rand}() < p_{rec}$}
        \State $v_{next} \gets \text{second node in } p^*$
        \State $v_{wrong} \gets$ random neighbor of $s$ s.t. $ v_{wrong} \neq v_{next}$
        \State $p_{rec} \gets \text{shortest path from } v_{wrong} \text{ to } g$
        \State $\mathcal{D} \gets \mathcal{D} \cup \{[s, g, (s) \oplus p_{rec}]\}$
    \EndIf
\EndFor
\Return $\mathcal{D}$
\end{algorithmic}
\end{algorithm}

\subsection{Path Generation and Augmentation}
We perform a 90/10 split on reachable node pairs for training and testing.
The test set contains only unique reachable node pairs.
For each training pair, we generate a diverse set of paths, including
(i) shortest and top-$K$ shortest paths,
(ii) detour paths obtained by temporarily removing intermediate nodes, and
(iii) recovery paths that simulate early suboptimal decisions followed by replanning.
The detailed generation procedure is summarized in Algorithm~\ref{alg:data_gen}.

\subsection{Incremental Representation}
To decouple transition logic from absolute node indices, we transform trajectories into an incremental format. Each path $(u_0, u_1, \dots, u_T)$ is mapped to a sequence $[S, G, \text{inc}_1, \dots, \text{inc}_T]$, where $\text{inc}_t = u_t - u_{t-1}$. 

The vocabulary $\mathcal{V}$ is partitioned into separate segments for nodes and relative increments to prevent ID collisions. This partitioning ensures the model learns to predict the next action as a mathematical offset from the current state, rather than simply memorizing global node co-occurrences or spatial relationships. All sequences are padded to a fixed block size for efficient batch training.

\subsection{Training Configurations}
We train a 6-layer, 6-head, 120-dimensional Transformer for 20,000 iteration to ensure convergence, providing sufficient model capacity to master the task. Optimization is performed using the AdamW optimizer ($\beta_1=0.9, \beta_2=0.95$) with a global batch size of 1,024 and weight decay of 0.1. The learning rate peaks at $5 \times 10^{-4}$ and follows a cosine decay schedule to $5 \times 10^{-5}$ after 1,000 warmup iterations. To maintain stability, we employ \texttt{bfloat16} mixed-precision training and gradient clipping at a threshold of 1.0.

\section{Computational Efficiency of LSE-MTP}
\label{sec:efficiency_analysis}

The procedural realization of the LSE-MTP training objective is summarized in Algorithm~\ref{alg:lse_mtp}. 
It shares the same structure as standard MTP, follows the standard Transformer forward pass, 
and introduces transition layers only during training to provide supervision for future steps.

To evaluate the practical scalability of our method, we measured the training throughput of LSE-MTP, NTP, and standard MTP on an NVIDIA GeForce RTX 3090 Ti. The results show that switching from NTP to multi-token prediction ($K=4$) leads to a 4.5\% increase in parameters and a 17\% drop in tokens per second, but adding the LSE constraints on standard MTP incurs almost no additional computational cost.

The training throughput of LSE-MTP remains comparable to that of standard MTP (around 425k tokens/s). This is because the latent consistency and semantic anchoring losses operate directly on the hidden representations through lightweight mean squared error (MSE) computations, which are far less expensive than the linear projections and vocabulary-wide classification heads required by standard MTP. Moreover, all auxiliary heads are discarded after training, so LSE-MTP introduces no additional latency during inference. These findings indicate that, once the infrastructure for multi-step supervision is in place, LSE-MTP provides an almost "cost-free" mechanism to bridge the gap between discrete token prediction and continuous world modeling.

\section{Further Analysis of LSE-MTP}
\label{app:further_analysis}

We generated new ER and USG graphs and reduced the training set to 50\% of reachable node pairs. Under the same training path generation setup, we evaluated all models on previously unseen path planning tasks.

\paragraph{Overall Navigation Accuracy.}
Table~\ref{tab:results-more} shows that with moderate LSE-MTP hyperparameters, performance consistently improves over the corresponding MTP models across planning horizons $K$. Setting $\lambda_s$ to zero noticeably degrades performance, underscoring the role of semantic alignment. As $K$ increases, navigation accuracy declines due to reliance on next-step predictions, and adding future prediction losses can weaken this next-step capability by forcing trade-offs between objectives.

\paragraph{Preserving Latent Space Discriminability.} Table~\ref{tab:detailed-similarity-appendix} shows that semantic anchoring ($\lambda_s$) preserves latent space discriminability. Without it, the latent space collapses and topologically distinct nodes become indistinguishable. Incorporating $\mathcal{L}_{semantic}$ keeps path alignment within a meaningful manifold and prevents representational collapse.

\paragraph{Suppressing Structural Hallucinations.} Table~\ref{tab:detailed-isp-appendix} shows that LSE reduces illegal shortcut paths (ISP). Standard MTP models with only distant supervision often skip intermediate steps in latent space. Anchoring predictions to intermediate hidden states enforces step-by-step transitions, reducing ISP and ensuring valid path connectivity.

\begin{algorithm*}
\caption{Latent Semantic Enhancement MTP}
\label{alg:lse_mtp}
\begin{algorithmic}[1]
\Require Input sequence $U = \{u_1, \dots, u_T\}$; Prediction horizon $K$; Weights $\lambda_l, \lambda_s$.
\Ensure Total training loss $\mathcal{L}_{total}$.
\State $\mathbf{H} \gets \text{BackboneTransformer}(U)$ \Comment{$\mathbf{H} = \{\mathbf{h}_1, \dots, \mathbf{h}_T\}$}
\State $\mathcal{L}_{NTP} \gets \text{CrossEntropy}(\text{Head}_{shared}(\mathbf{H}), U_{target})$
\State $\mathcal{L}_{MTP} \gets 0$
\For{$k = 2$ \textbf{to} $K$}
    \State $\hat{\mathbf{h}}_{k} \gets \text{LinearProj}_{k-1}(\mathbf{H})$ \Comment{Project current state to future latent}
    \State $\mathcal{L}_{CE}^{(k)} \gets \text{CrossEntropy}(\text{Head}_{shared}(\hat{\mathbf{h}}_{k}), U_{target+k-1})$
    
    \State \textit{// Latent Consistency: Align predicted latent with future ground-truth latent}
    \State $\mathcal{L}_{latent}^{(k)} \gets \text{MSE}(\hat{\mathbf{h}}_{k}[:T-k+1], \mathbf{H}[k:])$ 
    
    \State \textit{// Semantic Anchoring: Align predicted latent with target embeddings}
    \State $\mathbf{E}_{target} \gets \text{EmbeddingLayer}(U_{target+k-1}).\text{detach}()$
    \State $\mathcal{L}_{semantic}^{(k)} \gets \text{MSE}(\hat{\mathbf{h}}_{k}, \mathbf{E}_{target})$
    
    \State $\mathcal{L}_{MTP} \gets \mathcal{L}_{MTP} + \mathcal{L}_{CE}^{(k)} + \lambda_l \mathcal{L}_{latent}^{(k)} + \lambda_s \mathcal{L}_{semantic}^{(k)}$
\EndFor
\State $\mathcal{L}_{total} \gets \mathcal{L}_{NTP} + \mathcal{L}_{MTP}$
\State \Return $\mathcal{L}_{total}$
\end{algorithmic}
\end{algorithm*}

\begin{table*}[htbp]
\centering
\caption{Detailed navigation performance metrics across different horizons ($K$) and LSE hyperparameter configurations ($\lambda_l, \lambda_s$). Suc, Disc, and WT represent Success rate, Disconnection rate, and Wrong Target rate, respectively.}
\label{tab:results-more}
\small
\setlength{\tabcolsep}{10pt}
\begin{tabular}{l l c | ccc || ccc}
\toprule
\multirow{2}{*}{\textbf{Model}} & \multirow{2}{*}{\textbf{$K$}} & \multirow{2}{*}{\makecell{\textbf{Hyperparams} \\ $(\lambda_l, \lambda_s)$}} & \multicolumn{3}{c||}{\textbf{ER (Erdős--Rényi Graph)}} & \multicolumn{3}{c}{\textbf{USG (Urban Street Graph)}} \\
\cmidrule{4-6} \cmidrule{7-9}
& & & Suc $\uparrow$ & Disc $\downarrow$ & WT $\downarrow$ & Suc $\uparrow$ & Disc $\downarrow$ & WT $\downarrow$ \\
\midrule
NTP (1TP) & 1 & - & 91.80 & 5.95 & 2.25 & 96.22 & 2.57 & 1.21 \\
\midrule
MTP &  & - & 91.99 & 6.44 & \textbf{1.57} & 97.13 & 1.70 & 1.17 \\
LSE-MTP &  & (0.1, 0.1) & \textbf{92.68} & \textbf{5.43} & 1.89 & 97.98 & 1.31 & 0.71 \\
LSE-MTP & 2 & (0.3, 0.3) & 91.22 & 6.33 & 2.45 & \textbf{98.36} & \textbf{1.13} & \textbf{0.51} \\
LSE-MTP &  & (0.5, 0.5) & 91.05 & 6.66 & 2.30 & 97.92 & 1.33 & 0.75 \\
LSE-MTP &  & (0.3, 0)   & 85.32 & 11.34 & 3.35 & 97.33 & 2.00 & 0.67 \\
\midrule
MTP &  & - & 89.50 & 8.37 & 2.13 & 96.28 & 2.67 & 1.05 \\
LSE-MTP &  & (0.1, 0.1) & \textbf{90.62} & \textbf{7.41} & \textbf{1.98} & 97.19 & 1.58 & 1.23 \\
LSE-MTP & 3 & (0.3, 0.3) & 88.90 & 8.29 & 2.81 & \textbf{97.56} & \textbf{1.47} & \textbf{0.97} \\
LSE-MTP &  & (0.5, 0.5) & 87.91 & 9.12 & 2.96 & 97.13 & 2.06 & 0.81 \\
LSE-MTP &  & (0.3, 0)   & 88.36 & 8.87 & 2.77 & 97.03 & 2.18 & 0.79 \\
\midrule
MTP &  & - & 87.72 & \textbf{9.02} & 3.26 & 95.72 & 3.39 & 0.89 \\
LSE-MTP &  & (0.1, 0.1) & \textbf{87.81} & 9.53 & \textbf{2.66} & 97.17 & 2.00 & 0.83 \\
LSE-MTP & 4 & (0.3, 0.3) & 86.58 & 10.39 & 3.03 & \textbf{97.29} & \textbf{1.90} & \textbf{0.81} \\
LSE-MTP &  & (0.5, 0.5) & 85.12 & 11.16 & 3.71 & 96.75 & 2.04 & 1.21 \\
LSE-MTP &  & (0.3, 0)   & 85.27 & 11.44 & 3.28 & 96.34 & 2.75 & 0.91 \\
\bottomrule
\end{tabular}
\end{table*}

\begin{table*}[htbp]
\centering
\caption{Detailed representation similarity metrics across different horizons ($K$) and LSE hyperparameter configurations ($\lambda_l, \lambda_s$). $G$ and $P$ indicate Goal and current Position, with $=$ and $\neq$ representing identical or different conditions. Baseline refers to the $G\neq, P\neq$ condition.}
\label{tab:detailed-similarity-appendix}
\small
\setlength{\tabcolsep}{3.5pt} 
\begin{tabular}{l l c | cccc || cccc}
\toprule
\multirow{2}{*}{\textbf{Model}} & \multirow{2}{*}{\textbf{$K$}} & \multirow{2}{*}{\makecell{\textbf{Hyperparams} \\ $(\lambda_l, \lambda_s)$}} & \multicolumn{4}{c||}{\textbf{ER (Erdős--Rényi Graph)}} & \multicolumn{4}{c}{\textbf{USG (Urban Street Graph)}} \\
\cmidrule{4-7} \cmidrule{8-11}
& & & $G=, P=$ & $G=, P\neq$ & $G\neq, P=$ & Baseline & $G=, P=$ & $G=, P\neq$ & $G\neq, P=$ & Baseline \\
\midrule
NTP (1TP) & 1 & - & 0.267 & 0.108 & 0.091 & 0.016 & 0.248 & 0.112 & 0.118 & 0.051 \\
\midrule
MTP & \multirow{5}{*}{2} & - & 0.376 & 0.221 & 0.120 & 0.057 & 0.298 & 0.118 & 0.119 & 0.039 \\
LSE-MTP & & (0.1, 0.1) & 0.396 & 0.263 & 0.129 & 0.069 & 0.346 & 0.152 & 0.176 & 0.078 \\
LSE-MTP & & (0.3, 0.3) & 0.382 & 0.264 & 0.122 & 0.068 & 0.366 & 0.160 & 0.200 & 0.093 \\
LSE-MTP & & (0.5, 0.5) & 0.380 & 0.270 & 0.133 & 0.086 & 0.369 & 0.176 & 0.216 & 0.107 \\
LSE-MTP & & (0.3, 0)   & 0.676 & 0.612 & 0.475 & 0.438 & 0.814 & 0.724 & 0.736 & 0.681 \\
\midrule
MTP & \multirow{5}{*}{3} & - & 0.410 & 0.281 & 0.136 & 0.085 & 0.317 & 0.118 & 0.121 & 0.027 \\
LSE-MTP & & (0.1, 0.1) & 0.456 & 0.355 & 0.149 & 0.094 & 0.372 & 0.169 & 0.182 & 0.082 \\
LSE-MTP & & (0.3, 0.3) & 0.439 & 0.345 & 0.137 & 0.091 & 0.391 & 0.167 & 0.203 & 0.086 \\
LSE-MTP & & (0.5, 0.5) & 0.450 & 0.361 & 0.137 & 0.094 & 0.394 & 0.178 & 0.213 & 0.103 \\
LSE-MTP & & (0.3, 0)   & 0.760 & 0.724 & 0.557 & 0.534 & 0.830 & 0.732 & 0.748 & 0.690 \\
\midrule
MTP & \multirow{5}{*}{4} & - & 0.419 & 0.307 & 0.149 & 0.094 & 0.337 & 0.124 & 0.115 & 0.026 \\
LSE-MTP & & (0.1, 0.1) & 0.469 & 0.372 & 0.160 & 0.106 & 0.379 & 0.158 & 0.165 & 0.063 \\
LSE-MTP & & (0.3, 0.3) & 0.480 & 0.398 & 0.165 & 0.118 & 0.407 & 0.176 & 0.208 & 0.098 \\
LSE-MTP & & (0.5, 0.5) & 0.480 & 0.405 & 0.162 & 0.127 & 0.429 & 0.196 & 0.226 & 0.112 \\
LSE-MTP & & (0.3, 0)   & 0.812 & 0.787 & 0.650 & 0.632 & 0.825 & 0.725 & 0.742 & 0.686 \\
\bottomrule
\end{tabular}
\end{table*}

\begin{table*}[htbp]
\centering
\caption{Detailed next-step probability coupling and structural hallucinations across different horizons ($K$) and LSE hyperparameter configurations ($\lambda_l, \lambda_s$). ISP and Legal Prob denote illegal shortcut and valid action probabilities, respectively.}
\label{tab:detailed-isp-appendix}
\small
\setlength{\tabcolsep}{10pt}
\begin{tabular}{l l c | cc || cc}
\toprule
\multirow{2}{*}{\textbf{Model}} & \multirow{2}{*}{\textbf{$K$}} & \multirow{2}{*}{\makecell{\textbf{Hyperparams} \\ $(\lambda_l, \lambda_s)$}} & \multicolumn{2}{c||}{\textbf{ER (Erdős--Rényi Graph)}} & \multicolumn{2}{c}{\textbf{USG (Urban Street Graph)}} \\
\cmidrule{4-5} \cmidrule{6-7}
& & & ISP $\downarrow$ & Legal Prob $\uparrow$ & ISP $\downarrow$ & Legal Prob $\uparrow$ \\
\midrule
NTP (1TP) & 1 & - & $7.9 \times 10^{-5}$ & 0.989 & $2.2 \times 10^{-5}$ & 0.999 \\
\midrule
MTP & \multirow{5}{*}{2} & - & $1.22 \times 10^{-4}$ & 0.988 & $5.7 \times 10^{-5}$ & \textbf{0.998} \\
LSE-MTP & & (0.1, 0.1) & $\mathbf{4.6 \times 10^{-5}}$ & \textbf{0.990} & $\mathbf{2.5 \times 10^{-5}}$ & \textbf{0.998} \\
LSE-MTP & & (0.3, 0.3) & $5.2 \times 10^{-5}$ & \textbf{0.990} & $3.9 \times 10^{-5}$ & \textbf{0.998} \\
LSE-MTP & & (0.5, 0.5) & $6.2 \times 10^{-5}$ & 0.989 & $3.1 \times 10^{-5}$ & \textbf{0.998} \\
LSE-MTP & & (0.3, 0)   & $7.1 \times 10^{-5}$ & \textbf{0.990} & $2.7 \times 10^{-5}$ & \textbf{0.998} \\
\midrule
MTP & \multirow{5}{*}{3} & - & $1.10 \times 10^{-4}$ & 0.985 & $7.8 \times 10^{-5}$ & 0.995 \\
LSE-MTP & & (0.1, 0.1) & $9.1 \times 10^{-5}$ & \textbf{0.989} & $\mathbf{4.0 \times 10^{-5}}$ & \textbf{0.997} \\
LSE-MTP & & (0.3, 0.3) & $\mathbf{3.6 \times 10^{-5}}$ & \textbf{0.989} & $4.4 \times 10^{-5}$ & \textbf{0.997} \\
LSE-MTP & & (0.5, 0.5) & $1.06 \times 10^{-4}$ & 0.987 & $4.6 \times 10^{-5}$ & \textbf{0.997} \\
LSE-MTP & & (0.3, 0)   & $4.9 \times 10^{-5}$ & \textbf{0.989} & $4.7 \times 10^{-5}$ & \textbf{0.997} \\
\midrule
MTP & \multirow{5}{*}{4} & - & $1.57 \times 10^{-4}$ & 0.981 & $1.27 \times 10^{-4}$ & 0.991 \\
LSE-MTP & & (0.1, 0.1) & $\mathbf{1.03 \times 10^{-4}}$ & \textbf{0.986} & $\mathbf{5.2 \times 10^{-5}}$ & \textbf{0.996} \\
LSE-MTP & & (0.3, 0.3) & $1.85 \times 10^{-4}$ & 0.985 & $5.9 \times 10^{-5}$ & \textbf{0.996} \\
LSE-MTP & & (0.5, 0.5) & $1.14 \times 10^{-4}$ & 0.985 & $7.0 \times 10^{-5}$ & \textbf{0.996} \\
LSE-MTP & & (0.3, 0)   & $1.33 \times 10^{-4}$ & 0.984 & $8.5 \times 10^{-5}$ & 0.995 \\
\bottomrule
\end{tabular}
\end{table*}

\end{document}